\pdfoutput=1
\documentclass[11pt]{article}

\usepackage{emnlp2021}

\usepackage{times}
\usepackage{latexsym}

\usepackage[T1]{fontenc}
\usepackage[utf8]{inputenc}

\usepackage{microtype}
\usepackage{amsmath,amssymb,amsfonts}
\usepackage{algorithmic}
\usepackage{graphicx}
\usepackage{textcomp}
\usepackage{multirow}
\usepackage{adjustbox}

\newcommand\blankfootnote[1]{%
  \let\thefootnote\relax\footnotetext{#1}%
  \let\thefootnote\svthefootnote%
}

\title{Automated Generation of Accurate \& Fluent Medical X-ray Reports}

\author{Hoang T.N. Nguyen{*} \\
  University of Alberta  \\
  \And
  Dong Nie{*} \\
  University of North Carolina at Chapel Hill \\
 \AND
 Taivanbat Badamdorj\\
 University of Alberta\\
 \And
 Yujie Liu\\
 Guangzhou University of \\Chinese Medicine\\
 \And
 Yingying Zhu\\
 University of Texas \\at Arlington\\
 \AND
 Jason Truong\\
 University of Alberta\\
 \And
 Li Cheng\\
 University of Alberta\\
}
  
\begin{document}
\maketitle

\blankfootnote{Code is available at \url{https://github.com/ginobilinie/xray_report_generation}}

\begin{abstract}
Our paper focuses on automating the generation of medical reports from chest X-ray image inputs, a critical yet time-consuming task for radiologists. Unlike existing medical report generation efforts that tend to produce human-readable reports, we aim to generate medical reports that are both fluent and clinically accurate. 
This is achieved by our fully differentiable and end-to-end paradigm containing three complementary modules: taking the chest X-ray images and clinical history document of patients as inputs, our \textit{classification} module produces an internal checklist of disease-related topics, referred to as enriched disease embedding; the embedding representation is then passed to our transformer-based \textit{generator}, giving rise to the medical reports; meanwhile, our generator also produces the weighted embedding representation, which is fed to our \textit{interpreter} to ensure consistency with respect to disease-related topics. 
Our approach achieved promising results on commonly-used metrics concerning language fluency and clinical accuracy.
Moreover, noticeable performance gains are consistently observed when additional input information is available, such as the clinical document and extra scans of different views. 
% --------------
% Important Note:
% Avoid saying we outperform SOTAs. There might be recently published papers that do better and we are not aware of. 
% --------------
\end{abstract}

\section{Introduction}
\label{sec:introduction}
% --------------
% What is the problem? Why is it interesting and important?
% --------------
Medical reports are the primary medium, through which physicians communicate findings and diagnoses from the medical scans of patients. The process is usually laborious, where typing out a medical report takes on average five to ten minutes~\cite{MedRepACL18}; it could also be error-prone. 
This has led to a surging need for automated generation of medical reports, to assist radiologists and physicians in making rapid and meaningful diagnoses. 
Its potential efficiency and benefits could be enormous, especially during critical situations such as COVID or a similar pandemic. 
Clearly a successful medical report generation process is expected to possess two key properties: 
1) clinical accuracy, to properly and correctly describe the disease and related symptoms; 
2) language fluency, to produce realistic and human-readable text.

% --------------
% Why is it hard?
% --------------
Fueled by recent progresses in the closely related computer vision problem of image-based captioning~\cite{ST,Tran_2020_CVPR}, there have been a number of research efforts in medical report generation in recent years~\cite{MedRepACL18,MedRepACL19,MedRepNIPS18,MedRepAAAI19,xue2018multimodal,yuan2019automatic,tienet,yin2019,lovelace-mortazavi-2020-learning,Srinivasan_2020_ACCV}.
These methods often perform reasonably well in addressing the language fluency aspect; on the other hand, as is also evidenced in our empirical evaluation, their results are notably less satisfactory in terms of clinical accuracy.  
This we attribute to two reasons: one is closely tied to the textual characteristic of medical reports, which typically consists of many long sentences describing various disease related symptoms and related topics in precise and domain-specific terms. This clearly sets the medical report generation task apart from a typical image-to-text problem such as image-based captioning; 
another reason is related to the lack of full use of rich contextual information that encodes prior knowledge. 
These information include for example clinical document of the patient describing key clinical history and indication from doctors, and multiple scans from different 3D views -- information that are typically existed in abundance in practical scenarios, as in the standard X-ray benchmarks of Open-I~\cite{openi} and MIMIC-CXR~\cite{johnson2019mimic}.

\begin{figure*}[t]
    \centering
    \includegraphics[width=0.8\linewidth]{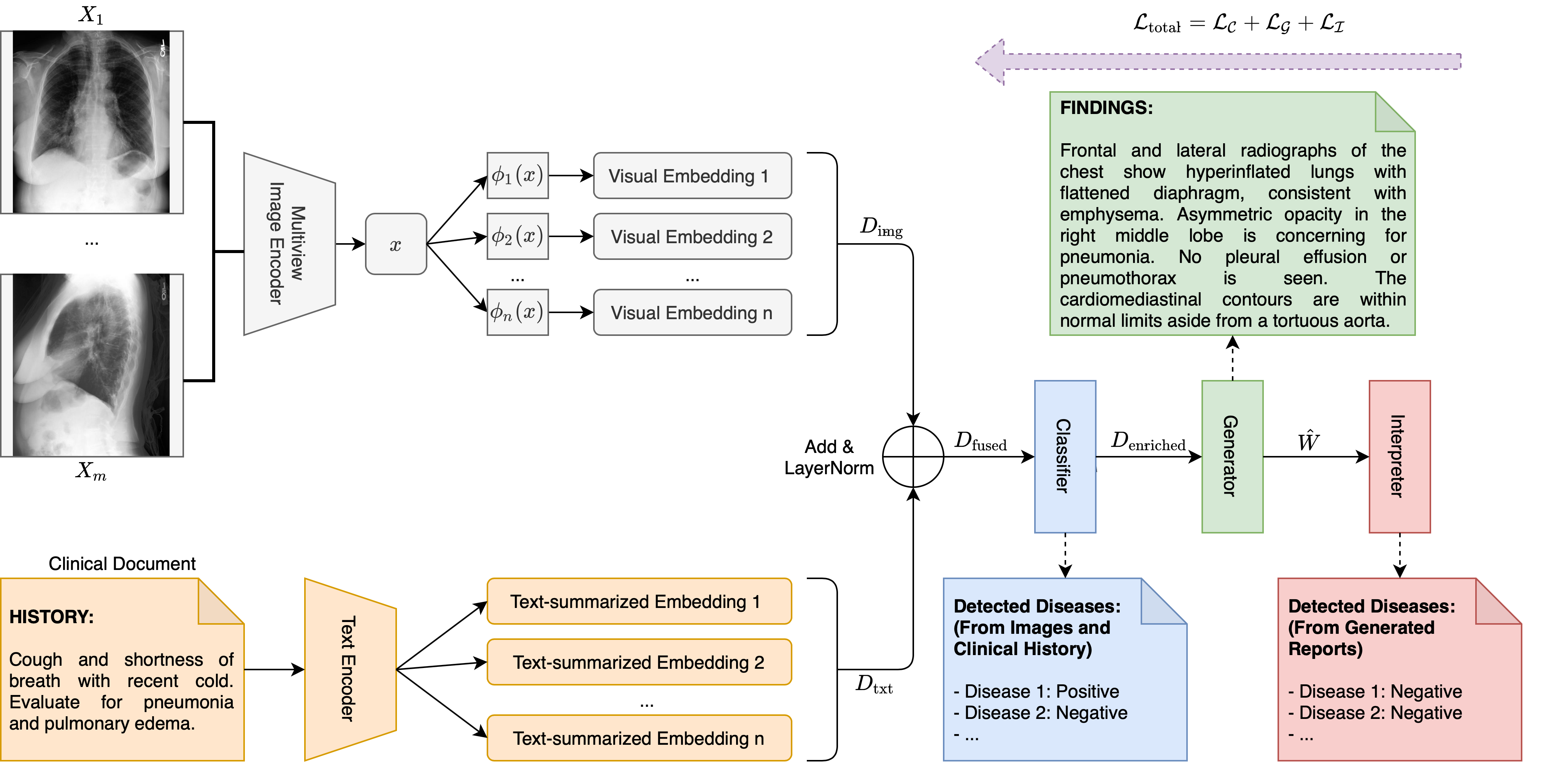}
    \caption{Our approach consists of three modules: a \textit{classifier} that reads chest X-ray images and clinical history to produce an internal checklist of disease-related topics; a transformer-based \textit{generator} to generate fluent text; an \textit{interpreter} to examine and fine-tune the generated text to be consistent with the disease-related topics.}
    \label{fig:classifier-generator-interpreter}
\end{figure*}

% --------------
% What are the key components of my approach and results? Also include any specific limitations.
% --------------
The aforementioned observations motivate us to propose a categorize-generate-interpret framework that places specific emphasis on clinical accuracy while maintaining adequate language fluency of the generated reports: a classifier module reads chest X-ray images (e.g., either single-view or multi-view images) and related documents to detect diseases and output enriched disease embedding; a transformer-based medical report generator; and a differentiable interpreter to evaluate and fine-tune the generated reports for factual correctness. 
% --------------
% Then have a final paragraph or subsection: "Summary of Contributions". It should list the major contributions in bullet form, mentioning in which sections they can be found. This material doubles as an outline of the rest of the paper, saving space and eliminating redundancy.
% --------------
% the use of visual transformer is relatively new here
The main contributions are two-fold: 
\begin{itemize}
    \item We propose a differentiable end-to-end approach consists of three modules (classifier-generator-interpreter) where the classifier module learns the disease feature representation via context modeling (section \ref{sec:context-modeling}) and disease-state aware mechanism (section \ref{sec:enriched-embedding}); the generator module transforms the disease embedding to medical report; the interpreter module reads and fine-tunes the generated reports, enhancing the consistency of the generated reports and the classifier's outputs.
    % \item Notably, unlike prior works \cite{MedRepACL19,MedRepNIPS18} that directly used visual features as inputs for text generation (i.e., learn features implicitly through report generation), our work first decouples the visual features into disease embedding and disease-states, then learns a state-aware disease embedding via the self-attention mechanism (i.e., learn features explicitly through disease classification). This significantly improve the report generation task (shown in Table~\ref{tab:openi-ablation}).
    \item Empirically our approach is demonstrated to be more competitive against many strong baselines over two widely-used benchmarks on an equal footing (i.e. without accessing to additional information). We also find that the clinical history of patients (prior-knowledge) play a vital role in improving the quality of the generated reports.
\end{itemize} 
% Please check these bullet points carefully! Reviewers will attack us based on these points

\section{Related Work}
\label{sec:related-work}
% It's critical to take a strong defensive stance about previous work right away. In this case Related Work can be either a subsection at the end of the Introduction, or its own Section 2.

\subsection{Image-based Captioning and Medical Report Generation}
Apart from some familiar topics such as disease detection~\cite{TMI_COVID,TMI_DEEP,lu2020multi,rajpurkar2017chexnet,lu2020muxconv,ranjan2018jointly} and lung segmentation~\cite{TMI_img2img}, the most related computer vision task is the emerging topic of image-based captioning, which aims at generating realistic sentences or topic-related paragraphs to summarize visual contents from images or videos~\cite{ST,SAT,goyal2017making,rennie2017self,huang2019attention,Feng_2019_CVPR,pei2019memory,Tran_2020_CVPR}.
Not surprisingly, the recent progresses in medical report generation~\cite{MedRepACL18,MedRepACL19,MedRepNIPS18,MedRepAAAI19,xue2018multimodal,yuan2019automatic,tienet,yin2019,lovelace-mortazavi-2020-learning,Srinivasan_2020_ACCV,zhang2020radiology,huang2021deepopht,gasimova2020spatial,singh2019chest,nishino2020reinforcement} have been particularly influenced by the successes in image-based captioning.

The work of~\cite{ST,SAT} is among the early approaches in medical report generation, where visual features are extracted by convolution neural networks (CNNs); they are subsequently fed into recurrent neural networks (RNNs) to generate textual descriptions. 
%Such models implicitly learn visual disease patterns via the image captioning process. However, these standard approaches have two major drawbacks. 
In remedying the issue of inaccurate textual descriptions, a secondary task is explicitly adopted by~\cite{MedRepACL18,Srinivasan_2020_ACCV} to select top-$k$ most likely diseases to gauge report generation. 
The methods of~\cite{MedRepACL19,MedRepNIPS18}, on the other hand, consider a reinforcement learning process to promote generating reports with correct contents.
It has been noted by~\cite{MedRepACL18,MedRepACL19,MedRepNIPS18} that traditional RNNs are not well suited in generating long sentences and paragraphs~\cite{vaswani2017attention,krause2017hierarchical}, which renders them insufficient in medical report generation task~\cite{MedRepACL18}. This issue is relieved by either conceiving hierarchical RNN architectures~\cite{krause2017hierarchical}~\cite{MedRepACL18,MedRepACL19,MedRepNIPS18,xue2018multimodal,yuan2019automatic,tienet,yin2019}, or resorting to alternative techniques including in particular the recently developed transformer architectures~\cite{vaswani2017attention}~\cite{Srinivasan_2020_ACCV,lovelace-mortazavi-2020-learning}. 
%Both architectures are well-suited for the task; however, 
%the work of \cite{Srinivasan_2020_ACCV,lovelace-mortazavi-2020-learning} shows that the transformer model has superior performance with faster training time compared to the RNN architectures. 

It is worth noting that most existing methods concentrates on the image-to-fluent-text aspect of the medical report generation problem; on the other hand, their results are considerably less well-versed at uncovering the intended disease and symptom related topics in the generated texts, the true gems where the physicians would base their decisions upon. 
%and fail to incorporate rich contextual information that experienced radiologists or doctors would consider. 
To alleviate this issue, a graph-based approach is considered in \cite{MedRepAAAI19}: it starts by compiling a list of common abnormalities, then transforms them into correlated disease graphs, and categorizes medical reports into templates for paraphrasing. Its practical performance is however less stellar, which may be credit to the fact that \cite{MedRepAAAI19} is fundamentally based on detecting abnormalities from medical images, thus may overlook other important information.

%In contrast, valuable raw data such as patients' symptoms, clinical history, and doctor's indication abundantly provided in most chest X-ray datasets such as Open-I~\cite{openi} and MIMIC-CXR~\cite{johnson2019mimic} is unused by all prior works. 
%Intuitively, a healthy patient without symptoms would be unlikely to visit a hospital for medical examination~\cite{taber2015people}. Therefore, the observed symptoms trigger the suspicion of medical doctors. As a result, a doctor's indication or reason for examination~\cite{wilcox2006written} is a vital question that doctors pose to radiologists given the observed clinical history and patients' symptoms. In other words, a radiology report must contain the answers to the doctors' questions. Therefore, the report generation task must accommodate medical doctors' practical needs and be re-formulated as a ``question-answering'' problem. 

%One reason is that medical images usually lack rich contextual information, which requires expert knowledge. Another reason is that each medical report is very domain-specific and precise and typically consists of many long sentences describing various diseases. 
%Therefore, the medical report generation task has three main challenges:
%1) describing diseases accurately (clinical coherence),
%2) producing realistic and human-readable sentences (language fluency), and
%3) producing complete descriptions in which all crucial information is included in the report (story completeness). % --> Only mention if we can measure. Try to come-up with a measurement.

\label{sec:method}
\begin{figure*}[t]
    \centering
    % angle=90,origin=c,
    \includegraphics[width=0.8\linewidth]{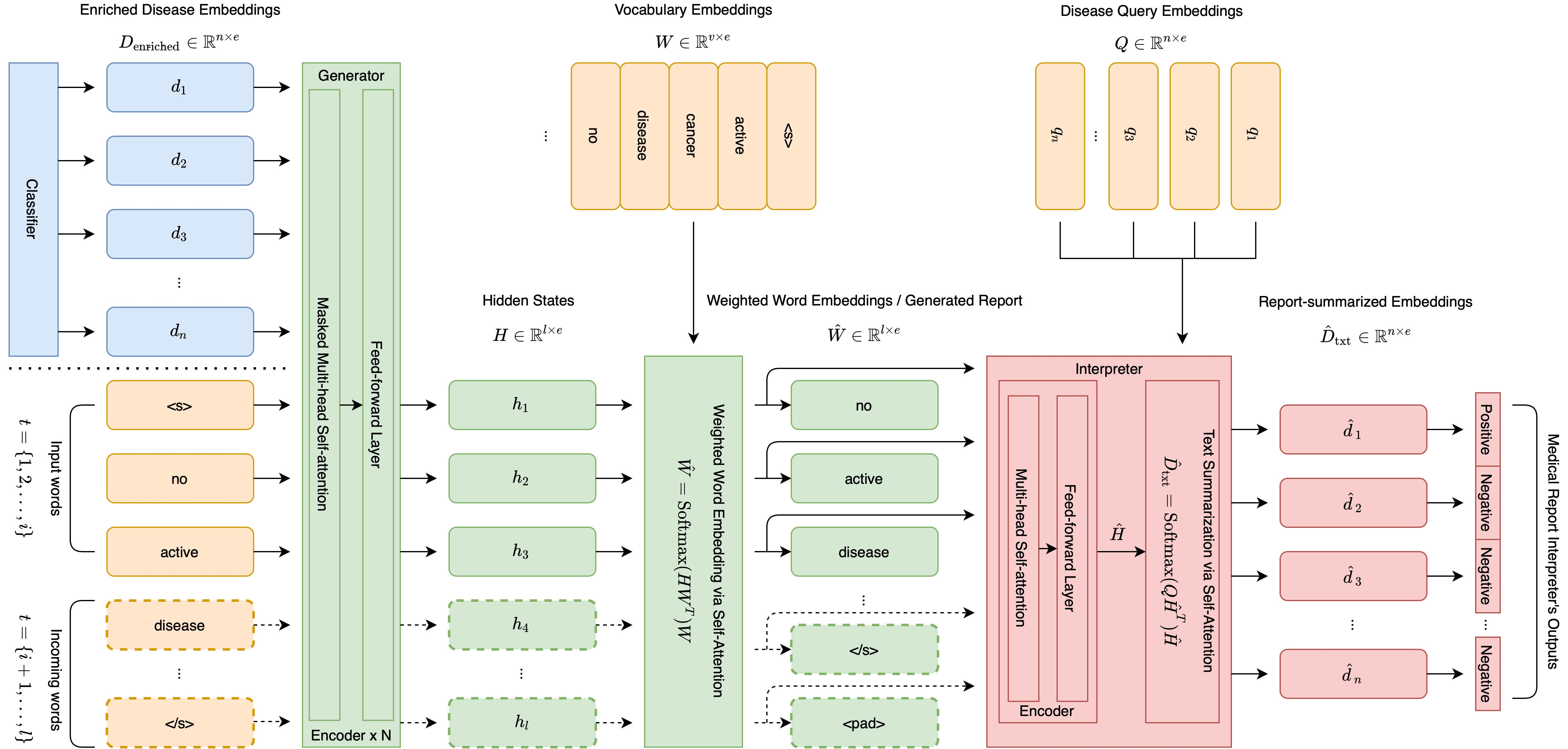}
    \caption{An example of our approach in action. The enriched disease embedding produced from the classification module are fed into the generation module as initial inputs. Then, at each time step, the hidden state $h_i$ is obtained and predicts the next output word. Finally, the interpretation module takes as input all predicted outputs $\hat{W}$ to predict a checklist of disease-related topics, which are to be gauged with the same topics output from the classification module for consistency verification. 
    %The interpreter's classification network is omitted for clarity.
    }
    \label{fig:generator-interpreter}
\end{figure*} % Color background to separate 3 components, add color similar to Fig.2., leveling the branches. Show another example because of the ___ (underscores). Change the disease term to "topics". Add a paragraph explain what is topic. Change the structure to to visual disease embedding 1, 2, 3, ... (avoid D). Similarly for detect diseases 1,2,3,... (avoid D)

\subsection{Transformers}
The transformer technique~\cite{vaswani2017attention} is first introduced in the context of machine translation with the purpose of expediting training and improving long-range dependency modeling. They are achieved by processing sequential data in parallel with an attention mechanism, consisting of a multi-head self-attention module and a feed-forward layer. By considering multi-head self-attention mechanisms, including e.g. a graph attention network \cite{velickovic2018graph}, recent transformer-based models have shown considerable advancement in many difficult tasks, such as image generation \cite{chen2020generative}, story generation \cite{radford2018improving}, question answering, and language inference \cite{devlin2018bert}. 

%For the sake of brevity, we fully adopted the Transformer Encoder model~\cite{vaswani2017attention} which consists of a multi-head self-attention module and a feed-forward layer. It models the relationship between words in the text via a self-attention mechanism. The usage of the Transformer's modules is discussed in later sections. Since the Transformer model is a new and fast-growing concept requiring in-depth explanation, we strongly encourage readers to refer to the original paper~\cite{vaswani2017attention}.
\subsection{CheXpert Labeler}
The CheXpert labeler~\cite{chexpert} is a rule-based system that extracts and classifies medical reports into 14 common diseases. Each disease label is either positive, negative, uncertain, or unmentioned. This is a crucial part in building large-scale chest X-ray datasets, such as~\cite{chexpert,johnson2019mimic}, where an alternative manual labeling process may take years of effort. It could also be used to evaluate the clinical accuracy of a generated medical report~\cite{pmlr-v106-liu19a}. 
Another important use of the CheXpert labeler is to facilitate the generation of medical reports. Since the rule-based CheXpert labeler is not differentiable, it is regarded as a score function estimator for reinforcement learning models~\cite{pmlr-v106-liu19a} to fine-tune the generated texts. However, the reinforcement learning methods are often computationally expensive and practically difficult to convergence. As an alternative, Lovelace et al.~\cite{lovelace-mortazavi-2020-learning} propose an attention LSTM model and fine-tune the generated report via a differentiable Gumbel random sampling trick, with promising results. 
%we propose a Transformer-based Interpreter model as a guidance system to fine-tune word representations via the self-attention mechanism. %The details are discussed in section~\ref{sec:interpreter}.

\section{Our Approach}
Our framework consists of a \textit{classification} module, a \textit{generation} module, and an \textit{interpretation} module, as illustrated in Fig.~\ref{fig:classifier-generator-interpreter}. 
The classification module reads multiple chest X-ray images and extracts the global visual feature representation via a multi-view image encoder. They are then disentangled into multiple low-dimensional visual embedding. Meanwhile, the text encoder reads clinical documents, including, e.g., doctor indication, and summarizes the content into text-summarized embedding. The visual and text-summarized embeddings are entangled via an ``add \& layerNorm'' operation to form contextualized embedding in terms of disease-related topics. 
The generation module takes our enriched disease embedding as initial input and generates text word-by-word, as shown in Fig.~\ref{fig:generator-interpreter}. Finally, the generated text is fed to the interpretation module for fine-tuning to align to the checklist of disease-related topics from the classification module. In what follows, we are to elaborate on these three modules in detail.

\subsection{The Classification Module}
\label{sec:classifier}

\subsubsection{Multi-view Image Encoder}
\label{sec:MVCNN}
For each medical study which consists of $m$ chest X-ray images $\{X_i\}_{i=1}^m$, we extract the corresponding latent features $\{x_i\}_{i=1}^m \in \mathbb{R}^c$, where $c$ is the number of features, via a shared DenseNet-121 image encoder~\cite{huang2017densely}. Then, the multi-view latent features $x \in \mathbb{R}^c$ can be obtained by max-pooling across the set of $m$ latent features $\{x_i\}_{i=1}^m$, as proposed in~\cite{su15mvcnn}. When $m=1$, the multi-view encoder boils down to a single-image encoder.

\subsubsection{Text Encoder}
\label{sec:TNN}
Let $T$ be a text document with length $l$ consisting of word embeddings $\{w_1, w_2,...,w_l\}$, where $w_i \in \mathbb{R}^e$ embodies the $i$-th word in the text and $e$ is the embedding dimension. We use the transformer encoder~\cite{vaswani2017attention} as our text feature extractor to retrieve a set of hidden states $H=\{h_1, h_2, ..., h_l\}$, where $h_i \in \mathbb{R}^e$ is the attended features of the $i$-th word to other words in the text,
\begin{equation} \label{eqn:encoder}
    h_i = \textrm{Encoder}(w_i|w_1,w_2,...,w_l).
\end{equation}

The entire document $T$ is then summarized by $Q=\{q_1,q_2,...,q_n\}$, representing $n$ disease-related topics (e.g., pneumonia or atelectasis) to be queried from the document. We refer to this retrieval process as \emph{text-summarized embedding} $D_\textrm{txt} \in \mathbb{R}^{n \times e}$,
\begin{equation} \label{eqn:dtxt}
    D_\textrm{txt} = \textrm{Softmax}\left(QH^\intercal\right) H.
\end{equation}
%Here the term $\textrm{Softmax}(QH^\intercal)$ shows word attention heat-map for the $n$ queried diseases in the document, as 
%\begin{equation} \label{eqn:dtxt_alpha}
%    \alpha = \textrm{Softmax}(QH^\intercal). % --> Confusing, merge to the above Equation.
%\end{equation}
Here matrix $Q \in \mathbb{R}^{n \times e}$ is formed by stacking the set of vectors $\{q_1,q_2,...,q_n\}$ where $q_i \in \mathbb{R}^e$ is randomly initialized, then learned via the attention process. Similarly, the matrix $H \in \mathbb{R}^{l \times e}$ is formed by $\{h_1, h_2, ..., h_l\}$ from Eq.~\eqref{eqn:encoder}. The term $\textrm{Softmax}(QH^\intercal)$ is the word attention heat-map for the $n$ queried diseases in the document.
%The text encoder is illustrated in Fig.~\ref{fig:classifier-generator-interpreter}.
%
The intuition here %behind Eq.~\eqref{eqn:dtxt} and Eq.~\eqref{eqn:dtxt_alpha} 
is for each disease (e.g., pneumonia) to be queried from the text document $T$. We only pay attention to the most relevant words (e.g., \textit{cough} or \textit{shortness of breath}) in the text that associates with that disease, 
%via Eq.~\eqref{eqn:dtxt_alpha}, 
also known as a vector similarity dot product. This way, the weighted sum of these words by Eq.~\eqref{eqn:dtxt} gives the feature that summarizes the document w.r.t. the queried disease. 

\subsubsection{Contextualized Disease Embedding}
\label{sec:context-modeling}
The latent visual features $x \in \mathbb{R}^{c}$ are subsequently decoupled into low-dimensional disease representations, as illustrated in Fig.~\ref{fig:classifier-generator-interpreter}. They are regarded as the \emph{visual embedding} $D_{\textrm{img}}\in \mathbb{R}^{n \times e}$, where each row is a vector $\phi_j(x) \in \mathbb{R}^e, j=1,\ldots,n$ defined as follows:
\begin{equation}
    \phi_j(x) = A_{j}^\intercal x + b_{j}. % Treat equation as a part of the equation, put comma, stop at the end.
\end{equation}
Here $A_j \in \mathbb{R}^{c \times e}$ and $b_j \in \mathbb{R}^{e}$ are learnable parameters of the $j$-th disease representation. 
$n$ is the number of disease representations, and $e$ is the embedding dimension. 
Now, together with the available clinical documents, the visual embedding $D_\textrm{img}$ and the text-summarized embedding $D_\textrm{txt}$ are entangled to form \emph{contextualized disease representations} $D_\textrm{fused} \in \mathbb{R}^{n \times e}$ as
\begin{equation}
    D_\textrm{fused} = \textrm{LayerNorm}(D_\textrm{img} + D_\textrm{txt}). % --> Explain layernorm in 1 sentence, and reference to the fig.1
\end{equation}

Intuitively, the entanglement of visual and textual information allows our model to mimic the hospital workflow, to screen the disease's visual representations conditioned on the patients' clinical history or doctors' indication. For example, the doctor's indication in Fig.~\ref{fig:classifier-generator-interpreter} shows \textit{cough} and \textit{shortness of breath} symptoms. It is reasonable for a medical doctor to request a follow-up check of the \textit{pneumonia} disease. As for the radiologists receiving the doctors' indication, they may prioritize diagnosing the presence of \textit{pneumonia} and related diseases based on X-ray scans and look for specific abnormalities. As empirically shown in Table~\ref{tab:openi-ablation}, the proposed contextualized disease representations bring a significant performance boost in the medical report generation task. 
Meanwhile, our current embedding is basically a plain mingling of heterogeneous sources of information such as disease type (i.e., disease name) and disease state (e.g., positive or negative).
As shown by the ablation study in Table~\ref{tab:openi-ablation}, this embedding by itself is insufficient for generating accurate medical reports. This leads us to conceive a follow-up enriched representation below. 

\subsubsection{Enriched Disease Embedding}
\label{sec:enriched-embedding}
The main idea behind \emph{enriched disease embedding} is to further encode informative attributes about disease states, such as \textit{positive}, \textit{negative}, \textit{uncertain}, or \textit{unmentioned}. 
% as inputs to our medical report generator. 
%
Formally, let $k$ be the number of states and $S \in \mathbb{R}^{k \times e}$ the state embedding. Then the confidence of classifying each disease into one of the $k$ disease states is
\begin{equation} 
\label{eqn:cls}
    p = \textrm{Softmax}(D_\textrm{fused}S^\intercal).
\end{equation}
$S \in \mathbb{R}^{k \times e}$ is randomly initialized, then learned via the classification of $D_\textrm{fused}$. $D_\textrm{fused}$ acts as features for the multi-label classification, and the classification loss is computed as
\begin{equation}
    \mathcal{L_C} = - \frac{1}{n} \sum_{i=1}^{n} \sum_{j=1}^{k} y_{ij} \log(p_{ij}),
\end{equation}
where $y_{ij} \in \{0,1\}$ and $p_{ij} \in [0,1]$ are the $j$-th ground-truth and predicted values for the disease $i$-th, respectively.
The state-aware embedding $D_\textrm{states} \in \mathbb{R}^{n \times e}$ are then computed as
\begin{equation}
    D_\textrm{states} = 
\left\{\begin{matrix}
yS, & \textrm{if training phase}\\ 
pS, & \textrm{otherwise}\\ 
\end{matrix}\right.
\end{equation}
$y \in \{0,1\}^{n \times k}$ is the one-hot ground-truth labels about the disease-related topics, whereas $p \in [0,1]^{n \times k}$ is the predicted values. During training, the ground-truth disease states facilitate our generator in describing the diseases \& related symptoms based on accurate information (teacher forcing). At test time, our generator then furnishes its recount based on the predicted states. %learned from the classification task.

Finally, the \emph{enriched disease embedding} $D_\textrm{enriched} \in \mathbb{R}^{n \times e}$ is the composition of state-aware disease embedding $D_\textrm{states}$ (i.e., good or bad), disease names $D_\textrm{topics}$ (i.e., which disease/topic), and the disease representations $D_\textrm{fused}$ (i.e., severity and details of the diseases),  
\begin{equation}
    D_\textrm{enriched} = D_\textrm{states} + D_\textrm{topics} + D_\textrm{fused}.
\end{equation}
Like the disease queries $Q$, $D_\textrm{topics} \in \mathbb{R}^{n \times e}$ is randomly initialized, representing diseases or topics to be generated. It is then learned in training through the medical report generation pipeline.
The enriched disease embedding provides explicit and precise disease descriptions, and endows our follow-up generation module with a powerful data representation. 
%as is also shown in Table~\ref{tab:openi-ablation}. 

\subsection{The Generation Module}
\label{sec:generator}
Our report generator is derived from the transformer encoder of~\cite{vaswani2017attention}. The network is formed by sandwiching \& stacking a masked multi-head self-attention component and a feed-forward layer being on top of each other for $N$ times, as illustrated in Fig.~\ref{fig:generator-interpreter}. The hidden state for each word position $h_i \in \mathbb{R}^e$ in the medical report is then computed based on previous words and disease embedding, as $D_{\textrm{enriched}}=\{d_i\}_{i=1}^n$, 
\begin{equation}
    h_i = \textrm{Encoder}(w_i|w_1,w_2,...,w_{i-1},d_1,d_2,...,d_n).
\end{equation}
This is followed by predicting future words based on the hidden states $H=\{h_i\}_{i=1}^l \in \mathbb{R}^{l \times e}$, as
\begin{equation}
    p_{\textrm{word}} = \textrm{Softmax}(HW^\intercal).
\end{equation}
Here $W \in \mathbb{R}^{v \times e}$ is the entire vocabulary embedding, $v$ the vocabulary size, and $l$ the document length. 
Let $p_{\textrm{word},ij}$ denote the confidence of selecting the $j$-th word in the vocabulary $W$ for the $i$-th position in the generated medical report. 
The generator loss is defined as a cross entropy of the ground-truth words $y_{\textrm{word}}$ and predicted words $p_{\textrm{word}}$, 
\begin{equation}
    \mathcal{L_G} = - \frac{1}{l} \sum_{i=1}^{l} \sum_{j=1}^{v} y_{\textrm{word},ij} \log(p_{\textrm{word},ij}).
\end{equation}

Finally, the \emph{weighted word embedding} $\hat{W} \in \mathbb{R}^{l \times e}$, also known as the \emph{generated report}, are:
\begin{equation}
    \hat{W} = p_{\textrm{word}}W.
\end{equation}
It is worth noting that this set-up facilitate the back-propagation of errors from the follow-up interpretation module.

\subsection{The Interpretation Module}
\label{sec:interpreter}
It is observed from empirical evaluations that the generated reports are often distorted in the process, such that they become inconsistent with the original output of the classification module -- the enriched disease embedding that encodes the disease and symptom related topics. 
Inspired by the CycleGAN idea of~\cite{ZhuEtAl:ICCV17}, we consider a fully differentiable network module to estimate the checklist of disease-related topics based on the generator's output, and to compare with the original output of the classification module. This provides a meaningful feedback loop to regulate the generated reports, which is used to fine-tune the generated report through the word representation outputs $\hat{W}$. 

Specifically, we build on top of the proposed text encoder (described in section~\ref{sec:TNN}) a classification network that classifies disease-related topics, as follows. First, the text encoder summarizes the current medical report $\hat{W}$, and outputs the report-summarized embedding of the queried diseases $Q$,
\begin{equation}
    \hat{D}_\textrm{txt} = \textrm{Softmax}(Q\hat{H}^\intercal)\hat{H} \in \mathbb{R}^{n \times e}.
\end{equation}
Here $\hat{H}$ is computed from the generated medical reports $\hat{W}$ using Eq.~\eqref{eqn:encoder}.
Second, each of the report-summarized embedding $\hat{d}_i \in \mathbb{R}^e$ (i.e., each row of the matrix $\hat{D}_\textrm{txt} \in \mathbb{R}^{n \times e}$) is classified into one of the $k$ disease-related states (i.e., positive or negative), as 
\begin{equation}
    p_{\textrm{int}} = \textrm{Softmax}(\hat{D}_\textrm{txt}S^\intercal) \in \mathbb{R}^{n \times k}.
\end{equation}
Finally, the interpreter is trained to minimize the subsequent multi-label classification loss, 
\begin{equation}
    \mathcal{L_I} = - \frac{1}{n} \sum_{i=1}^{n} \sum_{j=1}^{k} y_{ij} \log(p_{\textrm{int},ij}).
\end{equation}
here $y_{ij} \in \{0,1\}$ is the ground-truth disease label and $p_{\textrm{int},ij} \in [0,1]$ is the predicted disease label of the interpreter.

In fine-tuning the generated medical reports $\hat{W}$, all interpreter parameters are frozen, which acts as a guide to force the word representations $\hat{W}$ being close to what the interpreter has learned from the ground-truth medical reports. If the weighted word embedding $\hat{W}$ is different from the learned representation -- which leads to incorrect classification -- a large loss value will be imposed in the interpretation module. This thus forces the generator to move toward producing a correct word representation.

Collectively our model is trained in an end-to-end manner by jointly minimizing the total loss, 
\begin{equation}
    \mathcal{L}_\textrm{total} = \mathcal{L_C} + \mathcal{L_G} + \mathcal{L_I}.
\end{equation}

% \usepackage{multirow}
\begin{table*}[]
\begin{adjustbox}{width=2\columnwidth,center}
\centering
\tiny
\begin{tabular}{c|l|cccccc|c|c|c|c}
\hline
\multicolumn{1}{l|}{Datasets} &
  Methods &
  \multicolumn{1}{l}{B-1} &
  \multicolumn{1}{l}{B-2} &
  \multicolumn{1}{l}{B-3} &
  \multicolumn{1}{l}{B-4} &
  \multicolumn{1}{l}{MTR} &
  \multicolumn{1}{l|}{RG-L} &
%   \multicolumn{1}{l|}{CIDEr} &
  \multicolumn{1}{l|}{SV} &
  \multicolumn{1}{l|}{MV} &
  \multicolumn{1}{l|}{AI} &
  \multicolumn{1}{l}{FT} \\ \hline
\multirow{18}{*}{Open-I} &
  S\&T
  ~\cite{ST} 
  &
  0.316 &
  0.211 &
  0.140 &
  0.095 &
  0.159 &
  0.267 &
%   0.110 &
   x &
   &
   &
   \\
 &
  LRCN
  ~\cite{donahue2015long} 
  &
  0.369 &
  0.229 &
  0.149 &
  0.099 &
  0.155 &
  0.278 &
%   0.190 &
   x &
   &
   &
   \\
 &
  SA\&T
  ~\cite{SAT} 
  &
  0.399 &
  0.251 &
  0.168 &
  0.118 &
  0.167 &
  0.323 &
%   0.302 &
   x &
   &
   &
   \\
 &
  Att-RK
  ~\cite{you2016image} 
  &
  0.369 &
  0.226 &
  0.151 &
  0.108 &
  0.171 &
  0.323 &
%   0.155 &
   x &
   &
   &
   \\
 &
  HRNN
  ~\cite{8970668} 
  &
  0.445 &
  0.292 &
  0.201 &
  0.154 &
  0.175 &
  0.344 &
%   0.342 &
   x &
   &
   &
   \\
 &
  1-NN
  ~\cite{pmlr-v116-boag20a} 
  &
  0.232 &
  0.116 &
  0.051 &
  0.018 &
  N/A &
  0.201 &
%   0.728 &
   x &
   &
   &
   \\
 &
  TieNet
  ~\cite{tienet} 
  &
  0.330 &
  0.194 &
  0.124 &
  0.081 &
  N/A &
  0.311 &
%   1.334 &
   x &
   &
   &
   \\
 &
  Liu et. al.
  ~\cite{pmlr-v106-liu19a} 
  &
  0.359 &
  0.237 &
  0.164 &
  0.113 &
  N/A &
  0.354 &
%   \textbf{1.424} &
   x &
   &
   &
  x \\
 &
  CoAtt
  ~\cite{MedRepACL18} 
  &
  0.455 &
  0.288 &
  0.205 &
  0.154 &
  N/A &
  0.369 &
%   0.277 &
   x &
   &
   &
   \\
 &
  HRGR-Agent
  ~\cite{MedRepNIPS18} 
  &
  0.438 &
  0.298 &
  0.208 &
  0.151 &
  N/A &
  0.322 &
%   0.343 &
   &
  x &
   &
  x \\
 &
  KERP
  ~\cite{MedRepAAAI19} 
  &
  0.482 &
  0.325 &
  0.226 &
  0.162 &
  N/A &
  0.339 &
%   0.280 &
   &
  x &
  x &
   \\
 &
  ReinforcedTransformer
  ~\cite{10.1007/978-3-030-32692-0_77} 
  &
  0.350 &
  0.234 &
  0.143 &
  0.096 &
  N/A &
  N/A &
%   N/A &
   x &
   &
   &
  x \\
 &
  HRG-Transformer
  %~\cite{Srinivasan_2020_ACCV} 
  &
  0.464 &
  0.301 &
  0.212 &
  0.158 &
  N/A &
  N/A &
%   N/A &
   &
  x &
   &
   \\
 &
  SD\&C
  ~\cite{MedRepACL19} 
  &
  0.464 &
  0.301 &
  0.210 &
  0.154 &
  N/A &
  0.362 &
%   0.275 &
   x &
   &
   &
  x \\ \cline{2-12} 
 &
  Ours (SV) &
  0.463 &
  0.310 &
  0.215 &
  0.151 &
  0.186 &
  0.377 &
%   0.236 &
   x &
   &
   &
   \\
 &
  Ours (MV) &
  0.476 &
  0.324 &
  0.228 &
  0.164 &
  0.192 &
  0.379 &
%   0.249 &
   &
  x &
   &
   \\
 &
  Ours (MV+T) &
  0.485 &
  0.355 &
  0.273 &
  0.217 &
  0.205 &
  0.422 &
%   0.652 &
   &
  x &
  x &
   \\
 &
  Ours (MV+T+I) &
  \textbf{0.515} &
  \textbf{0.378} &
  \textbf{0.293} &
  \textbf{0.235} &
  \textbf{0.219} &
  \textbf{0.436} &
%   0.707 &
  &
  x &
  x &
  x \\ \hline %\hline
\multirow{10}{*}{MIMIC} &
  1-NN
  ~\cite{pmlr-v116-boag20a} 
  &
  0.367 &
  0.215 &
  0.138 &
  0.095 &
  0.139 &
  0.228 &
%   0.125 &
   x &
   &
   &
   \\
 &
  SA\&T
  ~\cite{SAT} 
  &
  0.370 &
  0.240 &
  0.170 &
  0.128 &
  0.141 &
  0.310 &
%   0.278 &
   x &
   &
   &
   \\
 &
  AdpAtt
  ~\cite{lu2017knowing} 
  &
  0.384 &
  0.251 &
  0.178 &
  0.134 &
  0.148 &
  0.314 &
%   0.299 &
   x &
   &
   &
   \\
 &
  Liu et. al.
  ~\cite{pmlr-v106-liu19a} 
  &
  0.313 &
  0.206 &
  0.146 &
  0.103 &
  N/A &
  0.306 &
%   \textbf{1.046} &
   x &
   &
   &
  x \\
 &
  Transformer
  ~\cite{vaswani2017attention} 
  &
  0.409 &
  0.268 &
  0.191 &
  0.144 &
  0.157 &
  0.318 &
%   0.318 &
   x &
   &
   &
   \\
 &
  GumbelTransformer
  ~\cite{lovelace-mortazavi-2020-learning} 
  &
  0.415 &
  0.272 &
  0.193 &
  0.146 &
  0.159 &
  0.318 &
%   0.316 &
   x &
   &
   &
  x \\ \cline{2-12} 
 &
  Ours (SV) &
  0.447 &
  0.290 &
  0.200 &
  0.144 &
  0.186 &
  0.317 &
%   0.224 &
   x &
   &
   &
   \\
 &
  Ours (MV) &
  0.451 &
  0.292 &
  0.201 &
  0.144 &
  0.185 &
  0.320 &
%   0.238 &
   &
  x &
   &
   \\
 &
  Ours (MV+T) &
  0.491 &
  0.357 &
  0.276 &
  0.223 &
  0.213 &
  0.389 &
%   0.537 &
  &
  x &
  x &
   \\
 &
  Ours (MV+T+I) &
  \textbf{0.495} &
  \textbf{0.360} &
  \textbf{0.278} &
  \textbf{0.224} &
  \textbf{0.222} &
  \textbf{0.390} &
%   0.502 &
   &
  x &
  x &
  x
\\
 \hline
\end{tabular}
\end{adjustbox}
\caption{
Quantitative comparison of our approach and many existing methods, evaluated under different setups of Single-view (SV), Multi-view (MV), w/ clinical text (T), and interpreter (I). For a fair comparison, all methods are categorized into the following four aspects: Single-View (SV), Multi-view (MV), Additional Information (AI), and Fine-tuning of the generated reports (FT). The best results are highlighted in \textbf{bold face}. Different language metrics are employed: BLEU-1 to BLEU-4 (B-1 to B-4), METEOR (MTR), and ROUGE-L (RG-L).
}
\label{tab:generator-performance}
\end{table*}

\section{Experiments}
\label{sec:experiments}
This section evaluates the medical report generation task on two fronts: the language performance and the clinical accuracy performance. 
Empirical evaluations are carried out on two widely-used chest X-ray datasets, MIMIC-CXR~\cite{johnson2019mimic} and Open-I~\cite{openi}.

\subsection{Datasets}
% Two widely-used benchmark datasets are used throughout our experiments. 
%For a fair comparison, the same train/val/test split is adopted by all the comparison methods in our experiments.

\subsubsection{MIMIC-CXR Dataset} 
The MIMIC-CXR dataset~\cite{johnson2019mimic} is a large-scale dataset with 227,835 medical reports of 65,379 patients, associated with 377,110 images from multiple views: anterior-posterior (AP), posterior-anterior (PA), lateral (LA). Each study comprises multiple sections, including \textit{comparison}, \textit{clinical history}, \textit{indication}, \textit{reasons for examination}, \textit{impressions}, and \textit{findings}. Here we utilize the multi-view images of AP/PA/LA views, and adopt as contextual information the concatenation of the \textit{clinical history}, \textit{reason for examination}, and \textit{indication} sections. For consistency, we follow the experimental set-up of~\cite{lovelace-mortazavi-2020-learning} to focus on generating text in the ``findings'' section as the corresponding medical report.
% In addition to the 14 disease labels extracted from the CheXpert labeler~\cite{chexpert}, we also obtain the top-100 high-frequency noun-phrases of the dataset as our additional disease-related topics using Spacy\footnote{Spacy is an NLP industrial open-source project: https://spacy.io}. In total, a list of 114 disease-related topics are acquired as the binary targets of the induced classification task. Each disease label is either \textit{positive} (including \textit{uncertain} or \textit{exist}) or \textit{negative} (including \textit{unmentioned} or \textit{non-exist}) Finally, 70\% of the medical studies are reserved for training, 10\% for validation, and 20\% for testing. We also ensure no patient overlap across the train and test sets to avoid data leakage.

\subsubsection{Open-I Dataset}
The Open-I dataset~\cite{openi} collected by the Indiana University hospital network contains 3,955 radiology studies that correspond to 7,470 frontal and lateral chest X-rays. Some radiology studies are associated with more than one chest X-ray image. Each study typically consists of \textit{impression}, \textit{findings}, \textit{comparison}, and \textit{indication} sections. 
Similar to the MIMIC-CXR dataset, we utilized both the multi-view chest X-ray images (frontal and lateral) and the \textit{indication} section as our contextual inputs. For generating medical reports, we follow the existing literature~\cite{MedRepACL18,Srinivasan_2020_ACCV} by concatenating the \textit{impression} and the \textit{findings} sections as the target output.
% As a preprocessing step, all tokens in the medical reports are converted to lowercase. 
% Since this dataset does not have any ground-truth disease labels, the 14 diseases extracted from the MIMIC-CXR dataset is used here, together with the top-100 high-frequency noun-phrases obtained in the Open-I dataset. This again forms the 114 disease-related topics as binary classification targets. Finally, the standard 70-10-20\% splits are utilized for training, validation, and testing purposes, respectively.

\subsubsection{Important Note} 
The implementation details, dataset splits, preprocessing steps, generated examples, and qualitative analysis are described in the supplementary materials.

\subsection{Experimental Results}

\subsubsection{Language Generation Performance}
A comprehensive quantitative comparison of our approach and many baselines 
%S\&T~\cite{ST}, LRCN~\cite{donahue2015long}, SA\&T~\cite{SAT}, Att-RK~\cite{you2016image}, HRNN~\cite{yin2019}, 1-NN~\cite{pmlr-v116-boag20a}, TieNet~\cite{tienet}, \cite{pmlr-v106-liu19a}, CoAtt~\cite{MedRepACL18}, HRGR-Agent~\cite{MedRepNIPS18}, KERP~\cite{MedRepAAAI19}, ReinforcedTransformer~\cite{10.1007/978-3-030-32692-0_77}, HRG-Transformer~\cite{Srinivasan_2020_ACCV}, SD\&C~\cite{MedRepACL19}, AdpAtt~\cite{lu2017knowing}, Transformer~\cite{vaswani2017attention}, and GumbelTransformer~\cite{lovelace-mortazavi-2020-learning} 
as shown in Table~\ref{tab:generator-performance} on the two benchmarks using the widely-used language evaluation metrics: BLEU-1 to BLEU-4~\cite{papineni2002bleu}, ROUGE-L~\cite{lin2004rouge}, and METEOR~\cite{banerjee2005meteor} scores.
%\footnote{Public API for language evaluation metrics: https://github.com/Maluuba/nlg-eval}
%  , and CIDEr~\cite{vedantam2015cider} scores.
%
Since all comparison methods have their own experiment setups, for a fair comparison, we further categorize these methods into four aspects: single-view (SV), multi-view (MV), accessing to additional information (AI) such as clinical document, and applying fine-tuning (FT) to the generated medical reports. Experiments in Table~\ref{tab:generator-performance} show that our models outperform the baselines in most language metrics.

With a single input X-ray image as the sole input, ours (SV) outperforms by a noticeable margin the best SOTA methods of CoAtt on Open-I and Transformer on MIMIC, respectively. This we mainly attribute to the utilization of the enriched disease embedding that explicitly incorporates the disease-related topics.
With multiple X-ray images as input, Ours (MV) again outperforms the best comparison methods of HRG-Transformer on Open-I. 
With multiple X-ray images and additional clinical document information as input, ours (MV+T) outperforms the comparison methods of KERP on Open-I.
Finally, with the complete contextual information available as input, ours (MV+T+I) outperforms all the comparison methods available in both Open-I and MIMIC datasets.

\subsubsection{Clinical Accuracy Performance}

\begin{table*}[]
\begin{adjustbox}{width=2\columnwidth,center}
\centering
\tiny
\begin{tabular}{c|l|c|cccc|cccc}
\hline
\multicolumn{3}{c|}{} & \multicolumn{4}{c|}{Macro scores} & \multicolumn{4}{c}{Micro scores} \\
Datasets                & Methods           & Acc.            & AUC      & F-1       & Prec.      & Rec.      & AUC      & F-1       & Prec.     & Rec.      \\ \hline
\multirow{5}{*}{Open-I} 
& 1-NN
~\cite{pmlr-v116-boag20a} 
& 0.911 & N/A & N/A & N/A & N/A & N/A & N/A & N/A & N/A \\
& S\&T
~\cite{ST} 
& 0.915 & N/A & N/A & N/A & N/A & N/A & N/A & N/A & N/A \\
& SA\&T
~\cite{SAT} 
& 0.908 & N/A & N/A & N/A & N/A & N/A & N/A & N/A & N/A \\
& TieNet
~\cite{tienet} 
& 0.902 & N/A & N/A & N/A & N/A & N/A & N/A & N/A & N/A \\
& Liu et. al.
~\cite{pmlr-v106-liu19a} 
& 0.918 & N/A & N/A & N/A & N/A & N/A & N/A & N/A & N/A \\ \cline{2-11}
& Ours (SV)         & 0.944          & 0.595          & 0.118          & 0.125          & 0.136          & 0.857          & 0.657          & 0.651          & \textbf{0.663} \\
                        & Ours (MV)         & 0.943          & 0.626          & 0.144          & 0.149          & 0.150          & \textbf{0.878} & 0.648          & 0.647          & 0.649          \\
                        & Ours (MV+T)       & \textbf{0.947} & 0.671          & 0.130          & \textbf{0.192} & 0.124          & 0.873          & \textbf{0.659} & \textbf{0.687} & 0.634          \\
                        & Ours (MV+T+I)     & 0.937          & \textbf{0.702} & \textbf{0.152} & 0.142          & \textbf{0.173} & 0.877          & 0.626          & 0.604          & 0.649          \\ \hline 
\multirow{9}{*}{MIMIC}  & 1-NN
~\cite{pmlr-v116-boag20a}
& N/A            & N/A            & 0.206          & 0.213          & 0.200          & N/A            & 0.335          & 0.346          & 0.324          \\
                        & SA\&T
                        ~\cite{SAT}
                        & N/A            & N/A            & 0.101          & 0.247          & 0.119          & N/A            & 0.282          & 0.364          & 0.230          \\
                        & AdpAtt
                        ~\cite{lu2017knowing}
                        & N/A            & N/A            & 0.163          & 0.341          & 0.166          & N/A            & 0.347          & 0.417          & 0.298          \\
                        & Liu et. al.
                        ~\cite{pmlr-v106-liu19a}
                        & 0.867            & N/A            & N/A          & 0.309          & 0.134          & N/A            & N/A          & \textbf{0.586}          & 0.237          \\
                        & Transformer
                        ~\cite{vaswani2017attention} 
                        & N/A            & N/A            & 0.214          & 0.327          & 0.204          & N/A            & 0.398          & 0.461          & 0.350          \\
                        & GumbelTransformer
                        ~\cite{lovelace-mortazavi-2020-learning} 
                        & N/A            & N/A            & 0.228          & 0.333          & 0.217          & N/A            & 0.411          & 0.475          & 0.361          \\ \cline{2-11} 
                        & Ours (SV)         & 0.877          & 0.743          & 0.342          & 0.357          & 0.347          & 0.857          & 0.530          & 0.533          & 0.528          \\
                        & Ours (MV)         & 0.880          & 0.752          & 0.347          & 0.385          & 0.347          & 0.862          & 0.533          & 0.545          & 0.522          \\
                        & Ours (MV+T)       & \textbf{0.890} & 0.778          & 0.407          & \textbf{0.448} & 0.399          & 0.872          & \textbf{0.578} & 0.583 & 0.574          \\
                        & Ours (MV+T+I)     & 0.887          & \textbf{0.784} & \textbf{0.412} & 0.432          & \textbf{0.418} & \textbf{0.874} & 0.576          & 0.567          & \textbf{0.585}
\\ \hline
\end{tabular}
\end{adjustbox}
\caption{
Quantitative comparison of clinical accuracy from the generated reports, evaluated on the 14 common CheXpert's diseases. The best results are highlighted in \textbf{bold face}.
}
\label{tab:generator-clinical}
\end{table*}

To evaluate the clinical accuracy of the generated reports, we use the LSTM CheXpert labeler~\cite{lovelace-mortazavi-2020-learning} as a universal measurement. We compare different methods based on accuracy, F-1, precision (prec.), and recall (rec.) metrics on 14 common diseases. Since there are 14 independent diseases, we also report the macro and micro scores. 
Intuitively, a high macro score means the detection of all 14 diseases is improved. Meanwhile, a high micro score implies the dominant diseases are improved (i.e., some diseases appear more frequently than others). As observed in Table~\ref{tab:generator-clinical}, our clinical performance increased significantly compared to the baselines in both macro and micro scores.

Among our ablation models in Table~\ref{tab:generator-clinical}, the precision and accuracy scores of our contextualized variant (MV+T) tend to be higher, whereas other scores are lower than the one with the interpreter (MV+T+I). This opposite behavior is due to the interpreter, which encourages detecting diseases, thus increases False Positives (FP). Note in the medical context, it is usually critically important to lower the False Negatives (FN) rate, thus a high recall score with a slight decrease in precision is more preferred.

\subsection{Ablation studies}
\label{sec:ablation-study}

\begin{table}[]
\centering
\tiny
\begin{tabular}{l|cccccc}
\hline
Methods                              & B-1 & B-2 & B-3 & B-4 & MTR & RG-L 
%& CIDEr 
\\ \hline
R w/o $D_{\textrm{states}}$        & 0.400  & 0.253  & 0.175  & 0.127  & 0.166  & 0.362   
%& 0.321 
\\
R w/o $D_{\textrm{topics}}$ & 0.453  & 0.300  & 0.206  & 0.142  & 0.183  & 0.366  
%& 0.204 
\\
R w/o $D_{\textrm{fused}}$ & 0.468  & 0.310  & 0.215  & 0.151  & 0.189  & 0.373   
%& 0.224 
\\
R with $D_{\textrm{enriched}}$ & 0.463  & 0.310  & 0.215  & 0.151  & 0.186  & 0.377   
%& 0.236 
\\
R + Interpreter         & 0.470  & 0.314  & 0.220  & 0.158  & 0.192  & 0.375   
%& 0.229 
\\ \hline
C w/o $D_{\textrm{states}}$ & 0.404  & 0.286  & 0.215  & 0.169  & 0.183  & 0.396   
%& 0.533 
\\
C w/o $D_{\textrm{topics}}$ & 0.474  & 0.329  & 0.244  & 0.187  & 0.194  & 0.401   
%& 0.472 
\\
C w/o $D_{\textrm{fused}}$ & 0.470  & 0.337  & 0.257  & 0.204  & 0.212  & 0.408   
%& 0.499 
\\
C with $D_{\textrm{enriched}}$ & 0.485  & 0.355  & 0.273  & 0.217  & 0.205  & 0.422   
%& 0.652 
\\
C + Interpreter & \textbf{0.515} & \textbf{0.378} & \textbf{0.293} & \textbf{0.235} & \textbf{0.219} & \textbf{0.436} 
%& \textbf{0.707}
\\ \hline
\end{tabular}
\caption{The table compares a regular image-to-text version (R) and a contextualized version (C) of our proposed method that utilizes clinical history on the Open-I dataset. For each version, we evaluate the importance of each component $D_\textrm{states}$, $D_\textrm{topics}$, and $D_\textrm{fused}$ in the proposed enriched disease embedding $D_\textrm{enriched}$ by removing one component at a time.}
\label{tab:openi-ablation}
\end{table}

\subsubsection{Enriched disease embedding}
We observe that the latent features $D_\textrm{fused}$ extracted from the classifier are insufficient to generate robust medical reports, as shown in Table~\ref{tab:openi-ablation}. Based on our human languages, a meaningful story needs three factors: the topic (i.e., what disease), the tone (i.e., is it negative or positive), and the details (i.e., the severity). However, there is no guarantee that the learned latent features $D_\textrm{fused}$ has all three required elements. On the other hand, with the the explicit representations (i.e., $D_\textrm{fused}$, $D_\textrm{topics}$, and $D_\textrm{states}$), all three factors are preserved. Therefore, the enriched disease embedding $D_\textrm{enriched}$ can generate precise and complete medical reports, leading to the language metrics' substantial improvement.

\subsubsection{Contextualized embedding}
Table~\ref{tab:openi-ablation} also shows that our proposed ``contextualized'' version can improve the language scores over the ``regular'' version, which reads only images.
Notably, the contextualized version is the entanglement of the chest X-ray images and the clinical history, which is crucial to improve the generated report's quality and accommodate doctors' practical needs. It mimics how radiologists receive requests from medical doctors and write reports to answer their questions.
Hence, the generated reports are believed to be more ``on point'' and receives higher language scores than the regular ``image-to-text'' setting.

% \subsubsection{Attention LSTM CheXpert Labeler vs Transformer-based Interpreter}
% In this part, we compare our proposed interpreter model with the LSTM attention proposed by~\cite{lovelace-mortazavi-2020-learning}. It is clear from Table~\ref{tab:interpreter-weight} that our proposed Interpreter is much more light-weight with about four times faster in training time. As shown in Table~\ref{tab:interpreter-performance}, the faster training time comes with a low disease detection score trade-off. However, as empirically shown in Tables~\ref{tab:generator-performance} and~\ref{tab:generator-clinical}, our Interpreter model's performance is still acceptable and helps achieving state-of-the-art results in other tasks.

% Note that because the LSTM CheXpert~\cite{lovelace-mortazavi-2020-learning} is more accurate than our Interpreter, and for consistent measurement, we measure the clinical accuracy using their model, as mentioned in the previous section.

\section{Conclusion and Outlook}
This paper introduces a novel three-module approach for generating medical reports from X-ray scans. Empirical findings demonstrated the superior performance of our approach over state-of-the-art methods on widely-used benchmarks under a range of evaluation metrics. %, where the comparisons are carried out on a fair ground. 
Moreover, our approach is flexible and can work with additional input information, where consistent performance gains are observed.  
%that our model boosted the generated reports' quality by leveraging contextual information and fine-tuning medical reports. 
For future work, we plan to apply our approach to related medical report generation tasks that go beyond X-rays. 
%and to consider disease localization and highlight disease terms automatically. 

\bibliographystyle{acl_natbib}
\bibliography{emnlp_main}

\begin{thebibliography}{52}
\expandafter\ifx\csname natexlab\endcsname\relax\def\natexlab#1{#1}\fi

\bibitem[{Banerjee and Lavie(2005)}]{banerjee2005meteor}
Satanjeev Banerjee and Alon Lavie. 2005.
\newblock Meteor: An automatic metric for mt evaluation with improved
  correlation with human judgments.
\newblock In \emph{Proceedings of the acl workshop on intrinsic and extrinsic
  evaluation measures for machine translation and/or summarization}, pages
  65--72.

\bibitem[{Boag et~al.(2020)Boag, Hsu, Mcdermott, Berner, Alesentzer, and
  Szolovits}]{pmlr-v116-boag20a}
William Boag, Tzu-Ming~Harry Hsu, Matthew Mcdermott, Gabriela Berner, Emily
  Alesentzer, and Peter Szolovits. 2020.
\newblock Baselines for chest x-ray report generation.
\newblock In \emph{NeurIPS Workshop on Machine Learning for Health}, pages
  126--140.

\bibitem[{Chen et~al.(2020)Chen, Radford, Child, Wu, Jun, Luan, and
  Sutskever}]{chen2020generative}
Mark Chen, Alec Radford, Rewon Child, Jeffrey Wu, Heewoo Jun, David Luan, and
  Ilya Sutskever. 2020.
\newblock Generative pretraining from pixels.
\newblock In \emph{International Conference on Machine Learning}, pages
  1691--1703. PMLR.

\bibitem[{Demner-Fushman et~al.(2016)Demner-Fushman, Kohli, Rosenman, Shooshan,
  Rodriguez, Antani, Thoma, and McDonald}]{openi}
Dina Demner-Fushman, Marc~D Kohli, Marc~B Rosenman, Sonya~E Shooshan, Laritza
  Rodriguez, Sameer Antani, George~R Thoma, and Clement~J McDonald. 2016.
\newblock Preparing a collection of radiology examinations for distribution and
  retrieval.
\newblock \emph{Journal of the American Medical Informatics Association},
  23(2):304--310.

\bibitem[{Devlin et~al.(2018)Devlin, Chang, Lee, and
  Toutanova}]{devlin2018bert}
Jacob Devlin, Ming-Wei Chang, Kenton Lee, and Kristina Toutanova. 2018.
\newblock Bert: Pre-training of deep bidirectional transformers for language
  understanding.
\newblock \emph{arXiv preprint arXiv:1810.04805}.

\bibitem[{Donahue et~al.(2015)Donahue, Anne~Hendricks, Guadarrama, Rohrbach,
  Venugopalan, Saenko, and Darrell}]{donahue2015long}
Jeffrey Donahue, Lisa Anne~Hendricks, Sergio Guadarrama, Marcus Rohrbach,
  Subhashini Venugopalan, Kate Saenko, and Trevor Darrell. 2015.
\newblock Long-term recurrent convolutional networks for visual recognition and
  description.
\newblock In \emph{Proceedings of the IEEE conference on computer vision and
  pattern recognition}, pages 2625--2634.

\bibitem[{Eslami et~al.(2020)Eslami, Tabarestani, Albarqouni, Adeli, Navab, and
  Adjouadi}]{TMI_img2img}
Mohammad Eslami, Solale Tabarestani, Shadi Albarqouni, Ehsan Adeli, Nassir
  Navab, and Malek Adjouadi. 2020.
\newblock \href {https://doi.org/10.1109/TMI.2020.2974159} {Image-to-images
  translation for multi-task organ segmentation and bone suppression in chest
  x-ray radiography}.
\newblock \emph{IEEE Transactions on Medical Imaging}, 39(7):2553--2565.

\bibitem[{Feng et~al.(2019)Feng, Ma, Liu, and Luo}]{Feng_2019_CVPR}
Yang Feng, Lin Ma, Wei Liu, and Jiebo Luo. 2019.
\newblock Unsupervised image captioning.
\newblock In \emph{Proceedings of the IEEE/CVF Conference on Computer Vision
  and Pattern Recognition}, pages 4125--4134.

\bibitem[{Gasimova et~al.(2020)Gasimova, Seegoolam, Chen, Bentley, and
  Rueckert}]{gasimova2020spatial}
Aydan Gasimova, Gavin Seegoolam, Liang Chen, Paul Bentley, and Daniel Rueckert.
  2020.
\newblock Spatial semantic-preserving latent space learning for accelerated dwi
  diagnostic report generation.
\newblock In \emph{International Conference on Medical Image Computing and
  Computer-Assisted Intervention}, pages 333--342. Springer.

\bibitem[{Goyal et~al.(2017)Goyal, Khot, Summers-Stay, Batra, and
  Parikh}]{goyal2017making}
Yash Goyal, Tejas Khot, Douglas Summers-Stay, Dhruv Batra, and Devi Parikh.
  2017.
\newblock Making the v in vqa matter: Elevating the role of image understanding
  in visual question answering.
\newblock In \emph{Proceedings of the IEEE Conference on Computer Vision and
  Pattern Recognition}, pages 6904--6913.

\bibitem[{Huang et~al.(2017)Huang, Liu, Van Der~Maaten, and
  Weinberger}]{huang2017densely}
Gao Huang, Zhuang Liu, Laurens Van Der~Maaten, and Kilian~Q Weinberger. 2017.
\newblock Densely connected convolutional networks.
\newblock In \emph{Proceedings of the IEEE conference on computer vision and
  pattern recognition}, pages 4700--4708.

\bibitem[{Huang et~al.(2021)Huang, Yang, Liu, Tian, Liu, Wu, Lin, Wang,
  Morikawa, Chang et~al.}]{huang2021deepopht}
Jia-Hong Huang, C-H~Huck Yang, Fangyu Liu, Meng Tian, Yi-Chieh Liu, Ting-Wei
  Wu, I~Lin, Kang Wang, Hiromasa Morikawa, Hernghua Chang, et~al. 2021.
\newblock Deepopht: medical report generation for retinal images via deep
  models and visual explanation.
\newblock In \emph{Proceedings of the IEEE/CVF winter conference on
  applications of computer vision}, pages 2442--2452.

\bibitem[{Huang et~al.(2019)Huang, Wang, Chen, and Wei}]{huang2019attention}
Lun Huang, Wenmin Wang, Jie Chen, and Xiao-Yong Wei. 2019.
\newblock Attention on attention for image captioning.
\newblock In \emph{Proceedings of the IEEE/CVF International Conference on
  Computer Vision}, pages 4634--4643.

\bibitem[{Irvin et~al.(2019)Irvin, Rajpurkar, Ko, Yu, Ciurea-Ilcus, Chute,
  Marklund, Haghgoo, Ball, Shpanskaya et~al.}]{chexpert}
Jeremy Irvin, Pranav Rajpurkar, Michael Ko, Yifan Yu, Silviana Ciurea-Ilcus,
  Chris Chute, Henrik Marklund, Behzad Haghgoo, Robyn Ball, Katie Shpanskaya,
  et~al. 2019.
\newblock Chexpert: A large chest radiograph dataset with uncertainty labels
  and expert comparison.
\newblock In \emph{Proceedings of the AAAI Conference on Artificial
  Intelligence}, volume~33, pages 590--597.

\bibitem[{Jing et~al.(2019)Jing, Wang, and Xing}]{MedRepACL19}
Baoyu Jing, Zeya Wang, and Eric Xing. 2019.
\newblock Show, describe and conclude: On exploiting the structure information
  of chest {X}-ray reports.
\newblock In \emph{Proceedings of the 57th Annual Meeting of the Association
  for Computational Linguistics}, pages 6570--6580.

\bibitem[{Jing et~al.(2018)Jing, Xie, and Xing}]{MedRepACL18}
Baoyu Jing, Pengtao Xie, and Eric Xing. 2018.
\newblock On the automatic generation of medical imaging reports.
\newblock In \emph{Proceedings of the 56th Annual Meeting of the Association
  for Computational Linguistics (Volume 1: Long Papers)}, pages 2577--2586.

\bibitem[{Johnson et~al.(2019)Johnson, Pollard, Berkowitz, Greenbaum, Lungren,
  Deng, Mark, and Horng}]{johnson2019mimic}
Alistair~EW Johnson, Tom~J Pollard, Seth~J Berkowitz, Nathaniel~R Greenbaum,
  Matthew~P Lungren, Chih-ying Deng, Roger~G Mark, and Steven Horng. 2019.
\newblock Mimic-cxr, a de-identified publicly available database of chest
  radiographs with free-text reports.
\newblock \emph{Scientific data}, 6(1):1--8.

\bibitem[{Krause et~al.(2017)Krause, Johnson, Krishna, and
  Fei-Fei}]{krause2017hierarchical}
Jonathan Krause, Justin Johnson, Ranjay Krishna, and Li~Fei-Fei. 2017.
\newblock A hierarchical approach for generating descriptive image paragraphs.
\newblock In \emph{Proceedings of the IEEE conference on computer vision and
  pattern recognition}, pages 317--325.

\bibitem[{Li et~al.(2018)Li, Liang, Hu, and Xing}]{MedRepNIPS18}
Christy~Y. Li, Xiaodan Liang, Zhiting Hu, and Eric~P. Xing. 2018.
\newblock Hybrid retrieval-generation reinforced agent for medical image report
  generation.
\newblock In \emph{NeurIPS}, pages 1537--1547.

\bibitem[{Li et~al.(2019)Li, Liang, Hu, and Xing}]{MedRepAAAI19}
Christy~Y. Li, Xiaodan Liang, Zhiting Hu, and Eric~P. Xing. 2019.
\newblock Knowledge-driven encode, retrieve, paraphrase for medical image
  report generation.
\newblock In \emph{AAAI}, pages 6666--6673.

\bibitem[{Lin(2004)}]{lin2004rouge}
Chin-Yew Lin. 2004.
\newblock Rouge: A package for automatic evaluation of summaries.
\newblock In \emph{Text summarization branches out}, pages 74--81.

\bibitem[{Liu et~al.(2019)Liu, Hsu, McDermott, Boag, Weng, Szolovits, and
  Ghassemi}]{pmlr-v106-liu19a}
Guanxiong Liu, Tzu-Ming~Harry Hsu, Matthew McDermott, Willie Boag, Wei-Hung
  Weng, Peter Szolovits, and Marzyeh Ghassemi. 2019.
\newblock Clinically accurate chest x-ray report generation.
\newblock In \emph{Machine Learning for Healthcare Conference}, pages 249--269.
  PMLR.

\bibitem[{Lovelace and Mortazavi(2020)}]{lovelace-mortazavi-2020-learning}
Justin Lovelace and Bobak Mortazavi. 2020.
\newblock Learning to generate clinically coherent chest x-ray reports.
\newblock In \emph{Proceedings of the 2020 Conference on Empirical Methods in
  Natural Language Processing: Findings}, pages 1235--1243.

\bibitem[{Lu et~al.(2017)Lu, Xiong, Parikh, and Socher}]{lu2017knowing}
Jiasen Lu, Caiming Xiong, Devi Parikh, and Richard Socher. 2017.
\newblock Knowing when to look: Adaptive attention via a visual sentinel for
  image captioning.
\newblock In \emph{Proceedings of the IEEE conference on computer vision and
  pattern recognition}, pages 375--383.

\bibitem[{Lu et~al.(2020{\natexlab{a}})Lu, Deb, and Boddeti}]{lu2020muxconv}
Zhichao Lu, Kalyanmoy Deb, and Vishnu~Naresh Boddeti. 2020{\natexlab{a}}.
\newblock Muxconv: Information multiplexing in convolutional neural networks.
\newblock In \emph{Proceedings of the IEEE/CVF Conference on Computer Vision
  and Pattern Recognition}, pages 12044--12053.

\bibitem[{Lu et~al.(2020{\natexlab{b}})Lu, Whalen, Dhebar, Deb, Goodman,
  Banzhaf, and Boddeti}]{lu2020multi}
Zhichao Lu, Ian Whalen, Yashesh Dhebar, Kalyanmoy Deb, Erik Goodman, Wolfgang
  Banzhaf, and Vishnu~Naresh Boddeti. 2020{\natexlab{b}}.
\newblock Multi-objective evolutionary design of deep convolutional neural
  networks for image classification.
\newblock \emph{IEEE Transactions on Evolutionary Computation}.

\bibitem[{Luo et~al.(2020)Luo, Yu, Chen, Liu, Wang, Xu, and Heng}]{TMI_DEEP}
Luyang Luo, Lequan Yu, Hao Chen, Quande Liu, Xi~Wang, Jiaqi Xu, and Pheng-Ann
  Heng. 2020.
\newblock \href {https://doi.org/10.1109/TMI.2020.3000949} {Deep mining
  external imperfect data for chest x-ray disease screening}.
\newblock \emph{IEEE Transactions on Medical Imaging}, 39(11):3583--3594.

\bibitem[{Nishino et~al.(2020)Nishino, Ozaki, Momoki, Taniguchi, Kano, Nakano,
  Tagawa, Taniguchi, Ohkuma, and Nakamura}]{nishino2020reinforcement}
Toru Nishino, Ryota Ozaki, Yohei Momoki, Tomoki Taniguchi, Ryuji Kano, Norihisa
  Nakano, Yuki Tagawa, Motoki Taniguchi, Tomoko Ohkuma, and Keigo Nakamura.
  2020.
\newblock Reinforcement learning with imbalanced dataset for data-to-text
  medical report generation.
\newblock In \emph{Proceedings of the 2020 Conference on Empirical Methods in
  Natural Language Processing: Findings}, pages 2223--2236.

\bibitem[{Oh et~al.(2020)Oh, Park, and Ye}]{TMI_COVID}
Yujin Oh, Sangjoon Park, and Jong~Chul Ye. 2020.
\newblock \href {https://doi.org/10.1109/TMI.2020.2993291} {Deep learning
  covid-19 features on cxr using limited training data sets}.
\newblock \emph{IEEE Transactions on Medical Imaging}, 39(8):2688--2700.

\bibitem[{Papineni et~al.(2002)}]{papineni2002bleu}
K.~Papineni et~al. 2002.
\newblock Bleu: a method for automatic evaluation of machine translation.
\newblock In \emph{ACL}, pages 311--318.

\bibitem[{Pei et~al.(2019)Pei, Zhang, Wang, Ke, Shen, and Tai}]{pei2019memory}
Wenjie Pei, Jiyuan Zhang, Xiangrong Wang, Lei Ke, Xiaoyong Shen, and Yu-Wing
  Tai. 2019.
\newblock Memory-attended recurrent network for video captioning.
\newblock In \emph{Proceedings of the IEEE/CVF Conference on Computer Vision
  and Pattern Recognition}, pages 8347--8356.

\bibitem[{Radford et~al.(2018)Radford, Narasimhan, Salimans, and
  Sutskever}]{radford2018improving}
Alec Radford, Karthik Narasimhan, Tim Salimans, and Ilya Sutskever. 2018.
\newblock Improving language understanding by generative pre-training.

\bibitem[{Rajpurkar et~al.(2017)Rajpurkar, Irvin, Zhu, Yang, Mehta, Duan, Ding,
  Bagul, Langlotz, Shpanskaya et~al.}]{rajpurkar2017chexnet}
Pranav Rajpurkar, Jeremy Irvin, Kaylie Zhu, Brandon Yang, Hershel Mehta, Tony
  Duan, Daisy Ding, Aarti Bagul, Curtis Langlotz, Katie Shpanskaya, et~al.
  2017.
\newblock Chexnet: Radiologist-level pneumonia detection on chest x-rays with
  deep learning.
\newblock \emph{arXiv preprint arXiv:1711.05225}.

\bibitem[{Ranjan et~al.(2018)Ranjan, Paul, Kapoor, Kar, Sethuraman, and
  Sheet}]{ranjan2018jointly}
Ekagra Ranjan, Soumava Paul, Siddharth Kapoor, Aupendu Kar, Ramanathan
  Sethuraman, and Debdoot Sheet. 2018.
\newblock Jointly learning convolutional representations to compress
  radiological images and classify thoracic diseases in the compressed domain.
\newblock In \emph{Proceedings of the 11th Indian Conference on Computer
  Vision, Graphics and Image Processing}, pages 1--8.

\bibitem[{Rennie et~al.(2017)Rennie, Marcheret, Mroueh, Ross, and
  Goel}]{rennie2017self}
Steven~J Rennie, Etienne Marcheret, Youssef Mroueh, Jerret Ross, and Vaibhava
  Goel. 2017.
\newblock Self-critical sequence training for image captioning.
\newblock In \emph{Proceedings of the IEEE Conference on Computer Vision and
  Pattern Recognition}, pages 7008--7024.

\bibitem[{Singh et~al.(2019)Singh, Karimi, Ho-Shon, and Hamey}]{singh2019chest}
Sonit Singh, Sarvnaz Karimi, Kevin Ho-Shon, and Len Hamey. 2019.
\newblock From chest x-rays to radiology reports: a multimodal machine learning
  approach.
\newblock In \emph{2019 Digital Image Computing: Techniques and Applications
  (DICTA)}, pages 1--8. IEEE.

\bibitem[{Srinivasan et~al.(2020)Srinivasan, Thapar, Bhavsar, and
  Nigam}]{Srinivasan_2020_ACCV}
Preethi Srinivasan, Daksh Thapar, Arnav Bhavsar, and Aditya Nigam. 2020.
\newblock Hierarchical x-ray report generation via pathology tags and multi
  head attention.
\newblock In \emph{Proceedings of the Asian Conference on Computer Vision}.

\bibitem[{Su et~al.(2015)Su, Maji, Kalogerakis, and Learned-Miller}]{su15mvcnn}
Hang Su, Subhransu Maji, Evangelos Kalogerakis, and Erik Learned-Miller. 2015.
\newblock Multi-view convolutional neural networks for 3d shape recognition.
\newblock In \emph{Proceedings of the IEEE international conference on computer
  vision}, pages 945--953.

\bibitem[{{Tran} et~al.(2020){Tran}, {Mathews}, and {Xie}}]{Tran_2020_CVPR}
A.~{Tran}, A.~{Mathews}, and L.~{Xie}. 2020.
\newblock \href {https://doi.org/10.1109/CVPR42600.2020.01305} {Transform and
  tell: Entity-aware news image captioning}.
\newblock In \emph{2020 IEEE/CVF Conference on Computer Vision and Pattern
  Recognition (CVPR)}, pages 13032--13042.

\bibitem[{Vaswani et~al.(2017)Vaswani, Shazeer, Parmar, Uszkoreit, Jones,
  Gomez, Kaiser, and Polosukhin}]{vaswani2017attention}
Ashish Vaswani, Noam Shazeer, Niki Parmar, Jakob Uszkoreit, Llion Jones,
  Aidan~N Gomez, Lukasz Kaiser, and Illia Polosukhin. 2017.
\newblock Attention is all you need.
\newblock \emph{arXiv preprint arXiv:1706.03762}.

\bibitem[{Veli{\v{c}}kovi{\'c} et~al.(2017)Veli{\v{c}}kovi{\'c}, Cucurull,
  Casanova, Romero, Lio, and Bengio}]{velickovic2018graph}
Petar Veli{\v{c}}kovi{\'c}, Guillem Cucurull, Arantxa Casanova, Adriana Romero,
  Pietro Lio, and Yoshua Bengio. 2017.
\newblock Graph attention networks.
\newblock \emph{arXiv preprint arXiv:1710.10903}.

\bibitem[{{Vinyals} et~al.(2015){Vinyals}, {Toshev}, {Bengio}, and
  {Erhan}}]{ST}
O.~{Vinyals}, A.~{Toshev}, S.~{Bengio}, and D.~{Erhan}. 2015.
\newblock \href {https://doi.org/10.1109/CVPR.2015.7298935} {Show and tell: A
  neural image caption generator}.
\newblock In \emph{2015 IEEE Conference on Computer Vision and Pattern
  Recognition (CVPR)}, pages 3156--3164.

\bibitem[{Wang et~al.(2018)Wang, Peng, Lu, Lu, and Summers}]{tienet}
Xiaosong Wang, Yifan Peng, Le~Lu, Zhiyong Lu, and Ronald~M Summers. 2018.
\newblock Tienet: Text-image embedding network for common thorax disease
  classification and reporting in chest x-rays.
\newblock In \emph{Proceedings of the IEEE conference on computer vision and
  pattern recognition}, pages 9049--9058.

\bibitem[{Xiong et~al.(2019)Xiong, Du, and Yan}]{10.1007/978-3-030-32692-0_77}
Yuxuan Xiong, Bo~Du, and Pingkun Yan. 2019.
\newblock Reinforced transformer for medical image captioning.
\newblock In \emph{International Workshop on Machine Learning in Medical
  Imaging}, pages 673--680. Springer.

\bibitem[{Xu et~al.(2015)Xu, Ba, Kiros, Cho, Courville, Salakhudinov, Zemel,
  and Bengio}]{SAT}
Kelvin Xu, Jimmy Ba, Ryan Kiros, Kyunghyun Cho, Aaron Courville, Ruslan
  Salakhudinov, Rich Zemel, and Yoshua Bengio. 2015.
\newblock Show, attend and tell: Neural image caption generation with visual
  attention.
\newblock In \emph{International conference on machine learning}, pages
  2048--2057. PMLR.

\bibitem[{Xue et~al.(2018)Xue, Xu, Long, Xue, Antani, Thoma, and
  Huang}]{xue2018multimodal}
Yuan Xue, Tao Xu, L~Rodney Long, Zhiyun Xue, Sameer Antani, George~R Thoma, and
  Xiaolei Huang. 2018.
\newblock Multimodal recurrent model with attention for automated radiology
  report generation.
\newblock In \emph{International Conference on Medical Image Computing and
  Computer-Assisted Intervention}, pages 457--466. Springer.

\bibitem[{{Yin} et~al.(2019)}]{yin2019}
C.~{Yin} et~al. 2019.
\newblock Automatic generation of medical imaging diagnostic report with
  hierarchical recurrent neural network.
\newblock In \emph{ICDM}, pages 728--737.

\bibitem[{Yin et~al.(2019)Yin, Qian, Wei, Li, Zhang, Li, and Zheng}]{8970668}
Changchang Yin, Buyue Qian, Jishang Wei, Xiaoyu Li, Xianli Zhang, Yang Li, and
  Qinghua Zheng. 2019.
\newblock Automatic generation of medical imaging diagnostic report with
  hierarchical recurrent neural network.
\newblock In \emph{2019 IEEE International Conference on Data Mining (ICDM)},
  pages 728--737. IEEE.

\bibitem[{You et~al.(2016)You, Jin, Wang, Fang, and Luo}]{you2016image}
Quanzeng You, Hailin Jin, Zhaowen Wang, Chen Fang, and Jiebo Luo. 2016.
\newblock Image captioning with semantic attention.
\newblock In \emph{Proceedings of the IEEE conference on computer vision and
  pattern recognition}, pages 4651--4659.

\bibitem[{Yuan et~al.(2019)Yuan, Liao, Luo, and Luo}]{yuan2019automatic}
Jianbo Yuan, Haofu Liao, Rui Luo, and Jiebo Luo. 2019.
\newblock Automatic radiology report generation based on multi-view image
  fusion and medical concept enrichment.
\newblock In \emph{International Conference on Medical Image Computing and
  Computer-Assisted Intervention}, pages 721--729. Springer.

\bibitem[{Zhang et~al.(2020)Zhang, Wang, Xu, Yu, Yuille, and
  Xu}]{zhang2020radiology}
Yixiao Zhang, Xiaosong Wang, Ziyue Xu, Qihang Yu, Alan Yuille, and Daguang Xu.
  2020.
\newblock When radiology report generation meets knowledge graph.
\newblock In \emph{Proceedings of the AAAI Conference on Artificial
  Intelligence}, volume~34, pages 12910--12917.

\bibitem[{Zhu et~al.(2017)Zhu, Park, Isola, and Efros}]{ZhuEtAl:ICCV17}
Jun-Yan Zhu, Taesung Park, Phillip Isola, and Alexei~A Efros. 2017.
\newblock Unpaired image-to-image translation using cycle-consistent
  adversarial networks.
\newblock In \emph{ICCV}.

\end{thebibliography}

\end{document}